# A SURVEY ON THE APPLICATION OF DATA SCIENCE AND ANALYTICS IN THE FIELD OF ORGANIZED SPORTS

Sachin Kumar S [1], Prithvi H V [2], C. Nandini [5]

[1] [2] **Final Year Students, Department of CSE, DSATM, Bengaluru, Karnataka, India**

[3] **Head of the CSE Department, DSATM, Bengaluru, Karnataka, India**

sachinsks1999@gmail.com[1]   hvprithvi09@gmail.com[2]   hodcse@dsatm.edu.in [5]

*Abstract—* The application of Data Science and Analytics to optimize or predict the outcomes is Ubiquitous in the Modern World. Data Science and Analytics have optimized almost every domain that exists in the market. In our survey we tend to focus mainly how the field of Analytics has been adopted in the field of sports, how it has contributed to the transformation of the game right from the assessment of on-field players and their selection to prediction of winning team and to the marketing of tickets and business aspects of big sports tournaments. We will present the analytical tools, algorithms and methodologies adopted in the field of Sports Analytics for different sports and also present our views on the same and we will also compare and contrast these existing approaches. By doing so, we will also present the best tools, algorithms and analytical methodologies to be considered by anyone who is looking to experiment with sports data and analyze various aspects of the game.

## I. INTRODUCTION

The term "Sports Analytics" gained widespread popularity in the field of Sports as well as in the field of Statistical Analysis following the release of a book titled "Moneyball: The Art of Winning an Unfair Game," by Michael Lewis [1], in the year 2003 and this popularity surmounted when the movie "Moneyball [2]" was released in the year 2011. In the story of "Moneyball," the Oakland Athletics General Manager, Billy Beane, who was also a former ballplayer, uses analytical methods to derive insights from the Player's historical data and makes pivotal decisions for his team, the Oakland A's. He relied intensely on the application of analytics to model an aggressive baseball team on a minimal budget, and his team set the world record for winning 20 consecutive games in the history of Baseball. After the success of Oakland Athletics in the game of Baseball by using Sabermetrics, an analytical approach adopted by Billy Beane to select low-cost players with high on-base percentage, many Baseball teams jumped-in on the strategy and later the use of Analytics to make decisions in the game of Sports was adopted in Basketball, American Football, Soccer, Tennis, Cricket and every other major highly popular organized sport. Sports Analytics came into existence with the experimentation of Statistical Analysis methods on the data of Baseball in the 1970s by the statisticians of the Society for American Baseball Research (SABR). Soon after its proven success by Billy Beane in the field of Baseball in optimizing player's and team's performance, people all around the world started experimenting with the datasets of Football, Hockey, Basketball, Soccer, Tennis, and Cricket. In the modern world, Sports Analytics is found to be used in almost every organized sport that is played. Today, we have Sports Analytics put into use in all primary sports right from Team-Selection and On-ground Decision making to business aspects of the sport. The development of this domain had its roots primarily from Statistics, Game Theory, and Decision Theory, and today, the field also uses Machine Learning and Modern Analytical Approaches to decisions on the team and the game itself.

Sports Analytics is the process that involves gathering all relevant, historical, and statistical data of players to develop a hypothesis based on statistical analysis and generate information that comes handy in making decisions for the team. There are two primary aspects when it comes to Sports Analytics, and these aspects apply to all kinds of Organized Games, and they are: "on-field analytics" and "off the field analytics." On the field analytics deals with the improvisation of the Player's performance or the team's performance on the field using insights from analytical reports. Off-the-field analytics deals with the business and market of sports. This comes useful when the team has to consider a particular budget for a tour and picking players who suit the total budget of the team for the tour. Sports Analytics is a more generalized term, and its application, although similar, has different approaches and outcomes in various sports. Usually, in every game, there is a blend of both the aspects in making decisions for a Team Selection and also for another decision making related to the team.

In the field of Sports Analytics concerning any sport, the most quintessential requirement to generate better analytics with the available empirical data on the competition is by asking the right questions and using the correct methodology to answer those questions. There can be more than one way to solve a problem, but the most desirable and efficient way always yields better results and better predictions. In the game of cricket, there is a plethora of Data available on the players and their strengths and weaknesses against rival teams and also their performances in various pitches around the world. To make a Team selection or decision for the team, asking the right questions is a start, and answering those questions with the right tools and most-proven methods is a necessity because the most reliable tools and techniques yield better predictions.

With the application of Sports Analytics to Sports we can get insights on the Sports Market, Assessment of Players and their Performances, Rank the different participating teams, Predict Player's score and team's score in a prospective game, Make Game Day Decisions to deliver a Win to the Team, Promote Team Brands and Team Products to improve the business and income of the Team, Optimize team's revenue and also Manage team's budget and finances. With such a wide range of applications, we have to consider the varied methodologies for each of these applications and select the best methods to maximize the outcome of the Application of Analytics. Literally, in the modern world, every decision in every major sport is made based on Analytics, but this knowledge is highly esoteric and not known to most of the people who aren't into the field of Analytics related to Sports and Sporting Events.

Building winning sports teams and successful sports businesses are more likely when data and models guide the decisions. Sports analytics is a source of competitive advantage. To apply the analytics feasibly to gain a competitive advantage in the field of sports, we should initially grasp the sport itself—the sports industry, the organized sports business, on-field factors and incidents, and courts of play. We have to envision and discern how to work with the data if sports—recognizing viable and verified data sources, gathering of relevant datasets to extract workable information, arranging, and modeling them to make it viable for investigation. Also, we need to learn from the data to consider how to assemble models from the information. Data cannot represent themselves. Informative pivotal forecasts don't emerge out of thin air. We must gain from information and build models that work.

## II. LITERATURE REVIEW

There is a plethora of literature available on Sports Analytics and Use of Analytics to make decisions in the sport of Baseball, Football, Basketball and Soccer, Tennis, and Hockey. Also, there exists a fewer number of research articles on the use of Analytics in the Sport of Cricket since 2008. But, the amount of literature on the use of Analytics in the sport of cricket is relatively low when compared to other games such as Baseball. In this section, we will give you an overview of the evolution of Sports Analytics from Baseball to Cricket, and different methods employed by different authors to make outcome predictions and team selections in the field of sports.

To introduce anyone to the vast field of Organized Sports Analytics, the best place to start with is the book "Moneyball: The Art of Winning an Unfair Game [1]", a book by Michael Lewis, published in 2003. The story of this book revolves around the General Manager of the Team, Oakland Athletics, Mr. Billy Beane, who set a world record by delivering 20 consecutive wins to the team with a minimal budget. The team was suffering from the shortage of funds, and Mr. Billy Beane adopted the principle of Sabermetrics, an Analytical approach to choose players for the team who cost relatively way less in price and had a more significant on-base percentage and slugging percentage. His choice of the on-base percentage and slugging percentage as better indicators of offensive success was by the use of rigorous statistical analysis. The Oakland A's' victory in the MLB Tournament in 2002 is a history that will last as long as the field of Sports and Sports Analytics holds water. James came up with the invented model the Sabermetrics, the year 1980 to sketch the science of analytics applied to the sport of baseball in honor of the Society for American Baseball Research (SABR), which was established in the year 1971. Yet, this analytical tool was not applied in practice because the Organization and members of the baseball team selection believed that a statistical tool could not surpass their years of experience. However, in the 2002 MLB Tournament, they were disproved by Billy Beane's use of Sabermetrics for selecting his Oakland A's Team. Beane's success was a result of Alderson's recommendation to Beane: they recognized which statistical tool most closely correlates the scoring of runs and other factors of the sport with the winning of the game, and at the same time the same factors that were undervalued by the rest of the baseball teams to make a selection decision. His analysis was that a player with an on-base percentage of .295 was paid around 4 million USD and a player who showcased an on-base percentage of .260 was paid only around 200 thousand USD. However, in the actual game setting, this overrated measure did not make much of a difference in delivering the most anticipated victory to the team all the time. Therefore, based on this analysis and the intuition, Billy Beane hired the player with an on-base percentage of .260 for a price that is a throw-away price when compared to buying a player with an on-base percentage of .295. Following the success of the Oakland A's in the MLB Tournament, analytics departments emerged in all the baseball team front offices.

Although the advent of Sports Analytics began with the game of Baseball in the early 1980s, the earliest organized sport was the Game of Cricket. We have reliable data on the scorecards of the games played since 1697 in the game of cricket. The record-keeping of games is what that gave rise to the introduction of Statistics and Analytics into the field of Sports. Of all the sorted out sports appreciated in America, it's maybe not astonishing that the one quintessential game to involve numbers as a prominent piece of its soul would be Baseball. Each occasion that happens in a ball game does as such with barely any undisclosed amounts: there are always nine defenders in the similar general territory, one-hitter, close to three base sprinters, close to 2 outs. There is no persistent whirlwind of action as in b-ball, soccer, or hockey; the occasions don't depend vigorously upon player arrangements as in football; there are no turnovers and no clock—each play is a discrete occasion. Thus, the advent of Sports Analytics took place with the game of Baseball. In spite of the availability of data for the game of cricket since 1697, the actual application of Analytics into organized sports happened with the game of Baseball. The most prominent reason for this unsung popularity of the sport, cricket in the field of Analytics was because it was not so popular in America. It was Americans that started applying statistics to sports data and Baseball was the most famous American Sport.

Spots that display the quintessential characteristic of placing two teams on a field of some size with a ball that is advanced towards a goal line has been around for as long as the human civilizations. The advent of the majestic game football is evolutionary and the fact that there exist different versions and variations of the game Football is perplexing. By the era of the 19th century, football was the game that was cherished by all the schools and universities across the world and prominently in the American, European and African Countries. With no authority governed by a revered central board at the time, each major school and/or university adopted a different variation of the sport depending on the surroundings of the university and the land that was available. Hence, we have the Association Football or the Soccer version of Football, and the rugby version of the football. In spite of these variations of the game, there were rudimentary data that was recorded judiciously since 19th century. Thus the application of Sports Analytics is somewhat viable in these games. In 1979, the revered organization, NCAA introduced an efficient metric called the "passer efficiency rating" as a reliable means to evaluate quarterbacks beyond touchdown passes and total yard and in the modern football games, all sorts of data are being recorded. Thus, the application of Analytics to these games are delivering more satisfactory results.

The third American sport that saw the application of Sports Analytics in recent years was Basketball. Like the game football, the game basketball was also cherished by colleges and imminent universities before the sport turned into an international organized sport with over a billion fans revering and following the sport in their routine lives. NBA Analysts are one of the highest-paid in the world and they run analytics on the game of Basketball to make decisions for the NBA Teams. In the information-rich era of the present-day modern world, the relationship between sports and numbers is closer than ever before. With the ease of access to all information publically available to anyone with internet access, it is easier than ever before to get the data about any organized sport and analyses can be run by anyone interested with a spreadsheet application to derive potential insights about the game from form of the game to the numbers that talk about the business of the sport. It is no longer as difficult as it was in the 1970s when people had to refer the newspaper to get statistical information about the sports and use a pocket calculator to perform statistical analysis on the aggregated information and report them with a manual typewriter.

## III. DATA SCIENCE TOOLS, ALGORITHMS, & METHODLOGIES ADOPTED IN THE FIELD OF SPORTS ANALYTICS

As mentioned earlier in the introduction section there are multiple applications of Data Science to the field of Sports for making different kinds of decisions and predictions for answering different set of questions. Each kind of decision making can be done in several ways but the goal of data science is to always adopt the method that yields higher number of successes over other methods. In this chapter we will discuss the different algorithms and methodologies adopted in the field of Sports Analytics specifically for Assessing Player's performance and predicting the team's outcome and Player's outcome in a prospective game. In all the literature that we have referred in the field of Sports Analytics the Methodology Adopted in every literature is the Method of Classification implemented by ML Classifier Models. But, the real difference in the output accuracy of these models lies within the choice of attributes to make the predictions. Therefore, it is true that the model contributes to the better predictions but the choice of attributes also influences the results in greater commensuration. When we try to achieve a replacement for a player/ to assess a player's performance in a new pitch where he hasn't played earlier, then, the K-Mean Clustering Model is the single most excellent model suggested by all the literature referenced in this research conduct. Yet, the choice of attributes plays a pivotal role in achieving higher results.

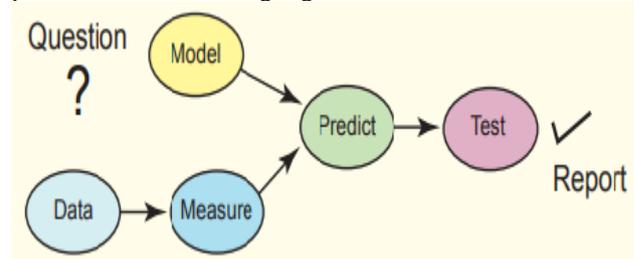

Figure 3.1: Blueprint for applying Data Science Techniques

### A. DATA SCIENCE METHODS

Data Scientists uncover the underlying facts and help to make data-informed decisions and this methodology is widely referenced as the fourth paradigm of science in the Modern World. There is a pre-defined template that the Data Scientists of the Industry adopts. This is a standard methodology of experimentation with data for analyzing the data and it is no different with sports data. The methodology is as listed as follows:

1. **Information search and selection.**
   Data Scientists review the literature and existing research methodologies in the field and learn what others have contributed in the past. Following the research of Literature, the next task is to look for relevant data sources and zero-down on the sources for analysis and modeling.

2. **Preparing text and data.**
   Text data is often unstructured data and sometimes it is partially structured data. It requires application of extensive data engineering and feature engineering principles to extract relevant information and prepare the data for analysis. Quantitative data are often messy data with a lot of unreliable data points or missing data points. They may require transformation and/or normalization prior to analysis. Data preparation is the most difficult and least appreciated task, but it is this step that makes all the difference in the final output of the data learning models.

3. **Looking at data.**
   After preparing the dataset that is reliable to conduct analysis, the immediate next task is to perform exploratory data analysis. This phase has two approaches and both are important to learn about the data. The two approaches are: Statistical EDA and visual EDA. EDA gives the information, summary and concept of the data

4. **Predicting how much.**
   To explain this task let's consider a viable example. When Data Scientists are asked to predict the quantitative value such as how many people might turn up to this upcoming tournament?, how many VIP seats will be sold?, how many scores will be scored by the home team? They base their predictions on the historical data and also run social analytics on the social media to gain insights on the trend and estimate a quantitative value with a percentage error.

5. **Predicting the outcomes as a definite yes/no.**
   Most business problems are identified as classification problems. Data Scientists use a variety of classification algorithms depending on the task at hand to predict the outcome as a definite Yes/ a No. We can illustrate this application with an example question: Will people buy tickets for this game?. Now that is a quintessential example of prediction task that predicts the outcome as a definite Yes/ No

6. **Testing out the analysis.**
   The task does not commence with the analysis and predictions made from the dataset. Whatever that is reported must be corroborated and to corroborate it, testing it out is essential. Data Scientists examine the outcome of their analytics with diagonistic tools to verify that the model works and it is reliable.

7. **Playing the game of what-if.**
   Data Scientists manipulate the depictor variables to know what happens to the predictions they've made. They play something that is what-if games in simulated marketplaces. They employ sensitivity testing or stress testing of mathematical programming models. The method of corroborating how change of values to variables affect the final outcomes, payoffs, and the predictions and assess the uncertainty present in the forecast models.

8. **Explaining it all.**
   Information extracted from the Big Data and presentation of insights and lerners that model the real world with data learning models, help us to understand the core functions of the domains that the data speaks about pertaining to the real world. Data Scientists turn what they have learned into an explanation that others can understand. It is the final task of a data scientist to present the results of the project in a clear and concise manner.

B. *LEARNING MODELS APPLICABLE TO THE DATA SETS OF ORGANIZED SPORTS*

By conducting extensive analysis on sports data sets, we may realize that learning the goal scorer's accuracy in the sport of football and basketball, or learning the favourable outcomes of a bowler is not the real deal. We may discover that the most valuable basketball player is not the one who scores the most points or makes the highest percentage of three-point shots, but the one who helps his teammates score. Through analytics, we discover many small findings of interest or curiosity in sports. But most importantly, we gain insight into sports business management and strategy. Just as we can view sports in a larger context as entertainment, we can view sports analytics within the larger context of data science. Information innovation experts convey information as information. Analysts and AI specialists center around models. Information researchers, expanding on an establishment of information and models, recount to a story that others can comprehend. They are worried about expectation, testing, and deciphering model outcomes, and if the information and models warrant, suggesting the board activity. Crafted by information science, delineated in figure 4.1, includes gaining from information and models and helping administrators settle on educated choices. We start with an inquiry. We end with a story, a report to the executives.

Prescient models come in two general structures, alluded to as grouping and relapse. Order concerns anticipating a future class or classification. Will the purchaser of sports purchase a pass to the game? If she purchases a pass to the game, will it be a ticket for a standard, liked, or a case seat. With characterization, we foresee purchaser decisions or discrete reactions. With relapse, then again, we anticipate a reaction with important size. What number of dollars will be spent on tickets? What amount of will be spent on concessions? What number of tickets will be offered to the game? What will be in the quantity of runs or focuses scored by the host group? What will be the correct defender's batting normal one year from now? What number of yards will the fullback run one week from now? What sort of compensation will a player get when he turns into a free specialist? A model is a portrayal of things, a rendering or depiction of the real world. An ordinary model in information science is an endeavor to relate one lot of factors to another. Constrained, uncertain, however valuable, a model causes us to understand the world. A model is something beyond talk since it depends on information.

Whether we want to assess a player and his performance, assess a team's performance, Rank Teams, Predict Scores of a Team or a Player, Predict the outcome of a Game or even manage finances of the team the one class of Machine Learning models that comes to our help is classification models. Of course, there are plentiful classification models to suit different kinds of classification problems and in this sub-section we will study a few Classification that has been used in the field of Sports Analytics.

The Figure 3.2 shows the most hackneyed methodology adopted by sports statisticians and analysts to model the sports data and perform analysis and build predictive models that predicts the win probabilities of the teams. This simple approach of considering the data of Visiting and Home teams as selected explanatory variable along with the game-day information to predict win probabilities is the stepping stone of sports predictive analytics.

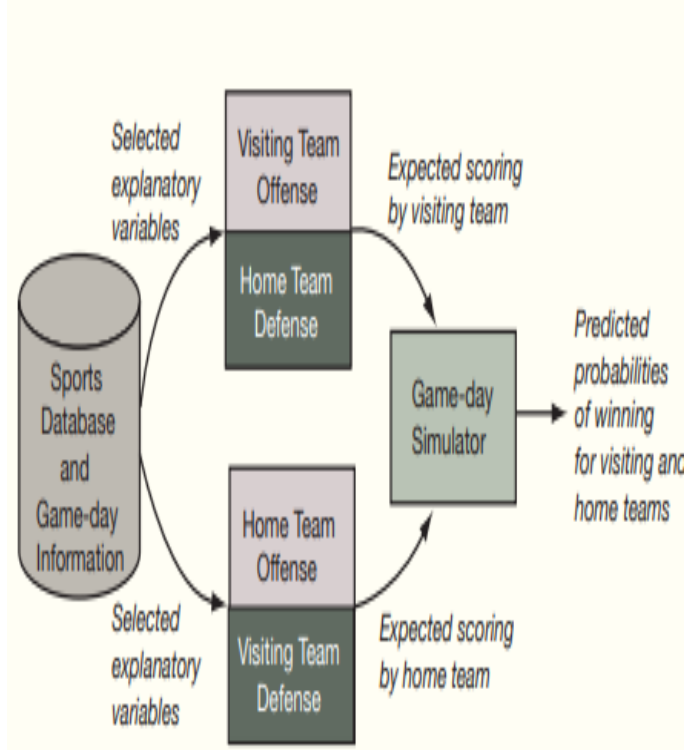

Figure 3.2: Predictive Modelling Framework For Team Sports

1. **NAIEVE BAYES CLASSIFIER**

A classifier is a machine learning model that is used to discriminate different objects based on certain features. Bayesian Network Algorithms are an esteemed category of Machine Learning Classifier Algorithms. They rely on Bayes Theorem which of conditional Probabilities. The assumption made here is that the predictors/features are independent. That is presence of one particular feature does not affect the other. Hence, it is called naive. Naïve Bayes classifier assumes that each attribute has its own individual effect on the class label, independent of the values of other attributes. This is called class-conditional independence. It is an excellent classifier algorithm that holds water in almost all real-world scenarios and datasets. But, the only limitation of this algorithm is that its output efficiency plummets when the dataset is a contiguous dataset and its' another prominent drawback is that the algorithm's accuracy plummets due to its nature of ignoring the information about conditional dependencies.

$$P(A|B) = \frac{P(B|A)P(A)}{P(B)}$$

Figure 3.3: Bayes Theorem

In the above figure 3.1 showing the Bayes theorem the term P(A|B) is the Posterior probability that we find using the theorem. This is nothing but the probability of occurance of A as the outcome by considering the Class Model, P(B|A) and the Class Prior Probability, P(A) in direct relation and the Predictor of Prior Probability, P(B) in the inverse relation with the prediction of the posterior probability. The classifier calculates P(A|B) for every class Ai for a given tuple B. It will then predict that B belongs to the class having the highest posterior probability, conditioned on B. That is B belongs to class Ai if and only if $P(A_i|B) > P(A_j|B)$ for $1 \leq j \leq m, j \neq i$.

2. **DECISION TREE CLASSIFIER**

Decision tree classifiers are non-parametric algorithms that build classification and regression models by using the principal concept of decision trees. The choice trees are worked such that Leaf hubs speak to the reliant class quality, and the root hub speaks to the autonomous information traits. To make an expectation, a choice tree is first produced and put away as the arrangement of rules to be utilized for deciding the class estimation of the new example. A representation of the Decision tree Algorithm regarding the agreement of Team's Win or Lose Prediction Based on the City, Venue, and Toss areas appeared underneath in figure 3.4.

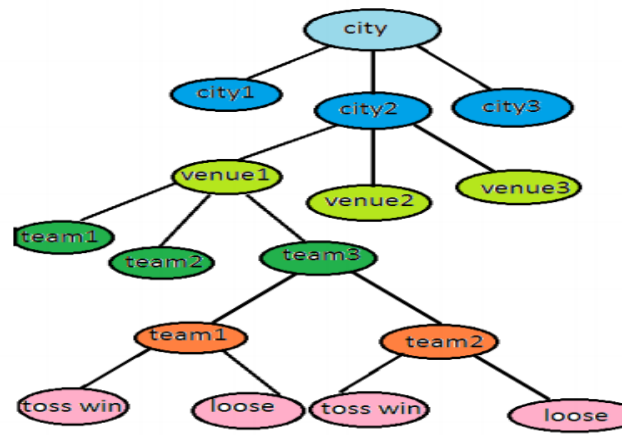

Figure 3.4: Decision Tree Illustration

There are various algorithms used in modelling the Decision Trees and they are:
- Gini Index
- Reduction in Variance
- Chi-Square
- ID3

Each algorithm has a notable uniqueness to it based on the selection bias of the algorithm. It is utterly necessary for us to use the right decision tree model to obtain the desired accuracy.

3. **RANDOM FOREST CLASSIFIER**

Random Forest is another type of tree-based classifier used for this thesis. It works by building multiple decision trees generated randomly from subsets of the training dataset. The classification of a new instance works by each decision tree predicts a class in which an example belongs then the category with most predictions is selected as the correct class of the situation. The benefits of this algorithm include robustness, and capability to handle large and high dimensional datasets appropriately. However, Random Forest has disadvantages as well. The first drawback is, Random Forest for regression problems is not as good as for classification problems. The second disadvantage is that the optimal number of features is not known intuitively, therefore trial and error required. Random forests are a set of decision trees where each tree is subject to an uncertain vector tested autonomously and with a similar dissemination of the considerable number of trees in the forest [18]. The algorithm generates several decision trees, creating a forest. Each decision tree is made by selecting random attributes at each node to determine the split [17]. It was Tim Kam Ho introduced the first method for random forests using a random subspace method in his paper [19]. Succeeding, Breiman Leo extended the algorithm in his paper [18], and this method was officially reffered as Random Forests.

### 4. SUPPORT VECTOR MACHINE

Support vector machine is another straightforward calculation that each AI master ought to have in his/her arms stockpile. Support vector machine is exceptionally favored by numerous individuals as it produces huge exactness with less calculation control. Support Vector Machine, curtailed as SVM can be utilized for both relapse and grouping errands. Be that as it may, it is broadly utilized in arrangement destinations. The goal of the help vector machine calculation is to discover a hyperplane in a N-dimensional space (N — the quantity of highlights) that unmistakably groups the information focuses.

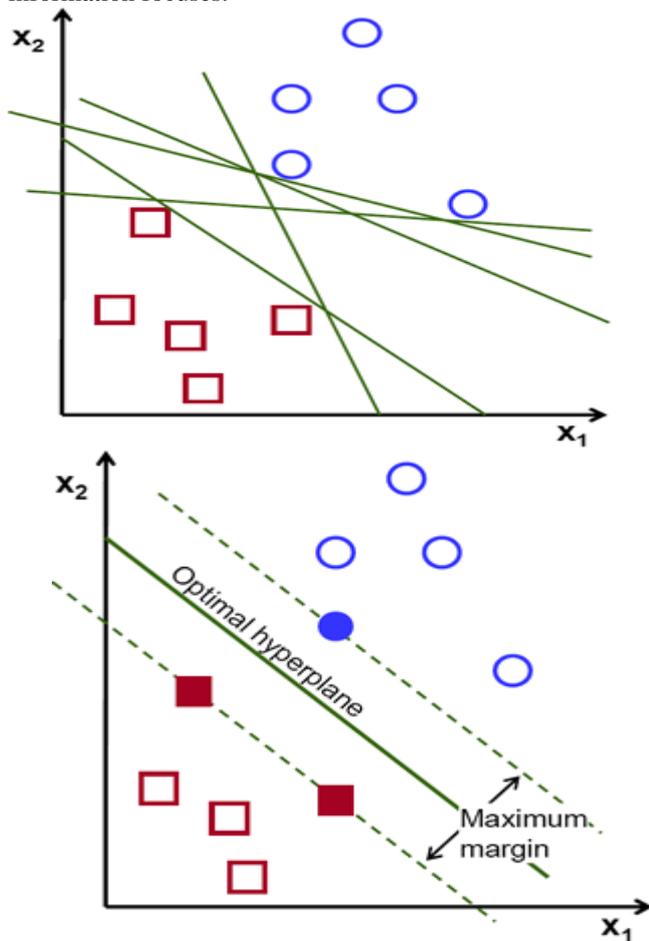

Figure 3.5: Possible Hyperplanes

Hyperplanes are choice limits that help arrange the information focuses. Information focuses falling on either side of the hyperplane can be ascribed to various classes. Likewise, the component of the hyperplane relies on the quantity of highlights. In the event that the quantity of info highlights is 2, at that point the hyperplane is only a line. In the event that the quantity of info highlights is 3, at that point the hyperplane turns into a two-dimensional plane. It gets hard to envision when the quantity of highlights surpasses 3. Bolster vectors are information indicates that are nearer the hyperplane and impact the position and direction of the hyperplane. Utilizing these help vectors, we amplify the edge of the classifier. Erasing the help vectors will change the situation of the hyperplane. These are the focuses that assist us with building our SVM. Hence, Support vector machine is an elegant and powerful algorithm.

### 5. K-MEANS ALGORITHM

To start with k-means algorithm, you first have to randomly initialize points called the cluster centroids (K). K-means is an iterative algorithm and it does two steps: 1. Cluster assignment 2. Move centroid step.

In Cluster task the calculation experiences every one of the information focuses and relying upon which bunch is nearer, It allots the information focuses to one of the three group centroids.

In the Move centroid step, K-implies moves the centroids to the normal of the focuses in a bunch. As it were, the calculation figures the normal of the considerable number of focuses in a bunch and moves the centroid to that normal area.

This procedure is rehashed until there is no adjustment in the bunches (or perhaps until some other halting condition is met). K is picked arbitrarily or by giving explicit beginning stages by the client.

The algorithm based on the input "K", tries to cluster the data points on the nearest centroid to the cluster and classifies creates K-Clusters. An Illustration of Clustering with K=3 is shown the fig 3.6.

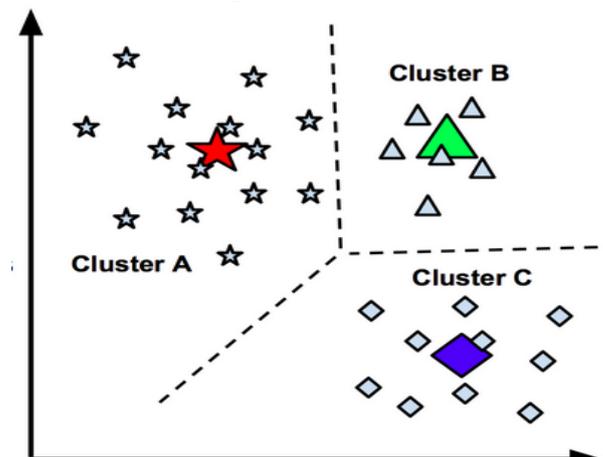

Figure 3.6: K-Means Clustering Algorithm with K=3

### 6. K-NEAREST NEIGHBOURS ALGORITHM
7. One of the most widely used algorithm in the field of Sports Analytics, the KNN Algorithm is a Supervised Learning class of ML Algorithms. It is a Non parametric as it does not make an assumption about the underlying data distribution pattern. It is used for both Classification and Regression and the choice of use entirely depends on the dateset. It's Utilization include similitude to foresee the group that the new point will fall into. K is a number used to distinguish comparable neighbors for the new information point. KNN takes K closest neighbors to choose where the new information point with have a place with. This choice depends on highlight similitude. Decision of K drastically affects the outcomes we acquire from KNN. Normally stepping through the exam set and plotting the exactness rate or F1 score against various estimations of K is the correct method for picking the incentive for K.

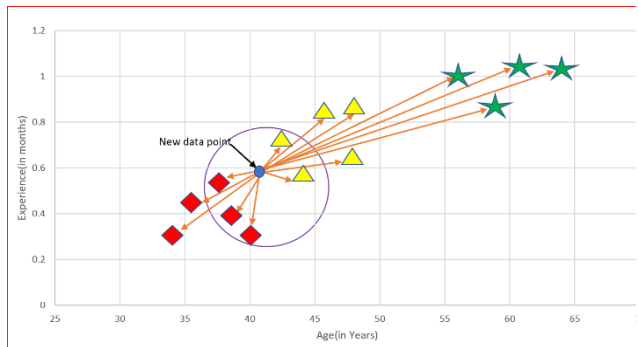

Figure 3.6: KNN Algorithm

## IV. Conclusion

Sports Analytics plays a major role in the field of organized sports and it has largely influenced the modern-day sports teams right from team selection to marketing of team brands. The application of Data Science to predict outcome of Sporting Events and to make a Winning team selection is largely identified as a Classification task. The grouping of players who has the similar gameplay is largely identified as a Clustering Task. In the application of Classification algorithms to answer relevant questions, not only the type of classification algorithm but also the attributes selection largely influences the outcome of the model and hence in the field of Sports Analytics it is most important to understand the influence of different attributes on the target data to predict certain outcomes.